\documentclass[letterpaper]{article} 
\usepackage{aaai2026}  
\usepackage{times}  
\usepackage{helvet}  
\usepackage{courier}  
\usepackage[hyphens]{url}  
\usepackage{graphicx} 
\usepackage{natbib}  
\usepackage{caption} 
\usepackage{algorithm}
\usepackage{algorithmic}

\usepackage{booktabs}
\usepackage{multirow}
\usepackage{amsmath}
\usepackage{amssymb}
\usepackage{pifont}
\usepackage{xcolor}
\usepackage[most]{tcolorbox}
\usepackage{tcolorbox}

\usepackage{newfloat}
\usepackage{listings}
\DeclareCaptionStyle{ruled}{labelfont=normalfont,labelsep=colon,strut=off} 
\floatstyle{ruled}
\newfloat{listing}{tb}{lst}{}
\floatname{listing}{Listing}
\title{SPEED-Q: Staged Processing with Enhanced Distillation towards Efficient Low-bit On-device VLM Quantization}
\author{
    Tianyu Guo, 
    Shanwei Zhao, 
    Shiai Zhu\thanks{Corresponding author.}, 
    Chenguang Ma
}
\affiliations{
    Ant Group \\
    \{guotianyu.gty, shanwei.zsw, shiai.zsa, chenguang.mcg\}@antgroup.com
}

\usepackage{bibentry}

\begin{document}

\maketitle

\begin{abstract}
Deploying Vision-Language Models (VLMs) on edge devices (e.g., smartphones and robots) is crucial for enabling low-latency and privacy-preserving intelligent applications. Given the resource constraints of these devices, quantization offers a promising solution by improving memory efficiency and reducing bandwidth requirements, thereby facilitating the deployment of VLMs. However, existing research has rarely explored aggressive quantization on VLMs, particularly for the models ranging from 1B to 2B parameters, which are more suitable for resource-constrained edge devices. In this paper, we propose \textbf{SPEED-Q}, a novel \textbf{S}taged \textbf{P}rocessing with \textbf{E}nhanc\textbf{E}d \textbf{D}istillation framework for VLM low-bit weight-only quantization that systematically addresses the following two critical obstacles: (1) significant discrepancies in quantization sensitivity between vision (ViT) and language (LLM) components in VLMs; (2) training instability arising from the reduced numerical precision inherent in low-bit quantization. In SPEED-Q, a staged sensitivity adaptive mechanism is introduced to effectively harmonize performance across different modalities. We further propose a distillation-enhanced quantization strategy to stabilize the training process and reduce data dependence. Together, SPEED-Q enables accurate, stable, and data-efficient quantization of complex VLMs. 
SPEED-Q is the first framework tailored for quantizing entire small-scale billion-parameter VLMs to low bits. 
Extensive experiments across multiple benchmarks demonstrate that SPEED-Q achieves up to $\mathbf{6\times}$ \textbf{higher accuracy} than existing quantization methods under 2-bit settings and consistently outperforms prior on-device VLMs under both 2-bit and 4-bit settings. Our code and models are available at \url{https://github.com/antgroup/SPEED-Q}.
\end{abstract}


\begin{figure}[t]
\centering
\includegraphics[width=0.95\columnwidth]{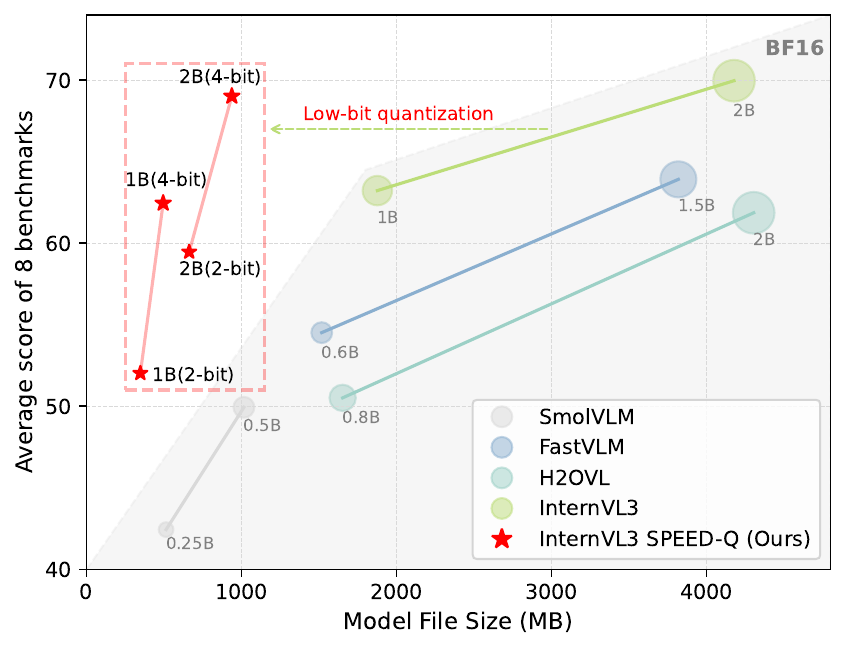}
\caption{Comparison with SOTA on-device VLMs. The proposed quantized version of InternVL3 outperforms the existing on-device VLMs with a smaller model file size.}
\label{fig:compare}
\end{figure}

\begin{figure}[t]
\centering
\includegraphics[width=0.95\columnwidth]{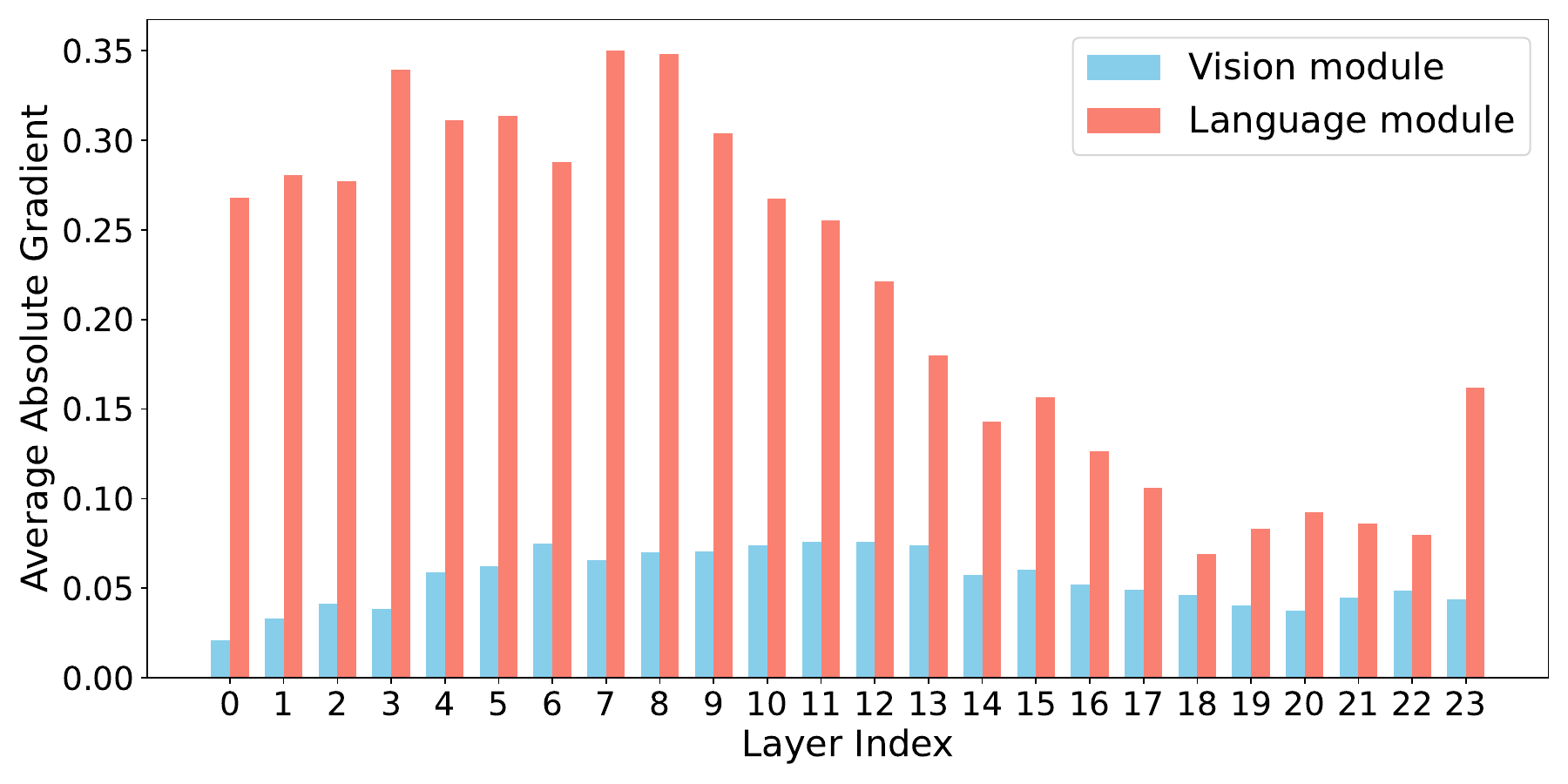}
\caption{The average absolute gradients of vision (ViT) and language (LLM) module in the InternVL2.5-1B quantization-aware training process.}
\label{fig:grad}
\end{figure}

\section{Introduction}

Vision-Language Models (VLMs) have achieved impressive performance across various applications, including visual question answering, robot navigation, and so on \cite{vlm-survey}.  However, the large size of VLMs presents significant challenges for deployment, particularly on resource-constrained edge devices. To facilitate on-device inference, a series of lightweight VLMs such as SmolVLM \cite{smolvlm}, FastVLM \cite{fastvlm}, and H2OVL \cite{h2ovl} have been proposed, aiming to reduce storage and memory costs. However, these models typically experience accuracy degradation due to the effects of scaling laws~\cite{scaling-law}.

Recently, several advanced large model series have released relatively small variants ranging from 1B to 2B parameters, such as InternVL3-2B \cite{internvl} and Qwen2-VL-2B \cite{wang2024qwen2}, which achieve performance comparable to earlier larger VLMs. However, their model size still poses significant challenges for on-device applications. For example, InternVL3-2B consumes over 4 GB of memory, pushing memory usage close to system limits and causing application instability.

Starting from the original carefully-designed models, quantization approaches reduce the precision of weights or activations from FP16 to lower bits, and substantially decrease computational resource demand while preserving model capacity. Thus, it has been an effective way to achieve better accuracy–efficiency trade-offs. Most previous works \cite{yu2025mquant, mbq, qslaw} focus on 8/4-bit quantization and have restricted quantization to language modules, leaving the problem of unified quantization for both vision and language components largely unexplored. This limitation poses a significant challenge for deploying VLMs with 1–2B parameters, as the visual module accounts for a substantial portion of the parameters. Taking InternVL3-1B as an example, the ViT encoder contains 304M out of its 938M total parameters. Therefore, efficient deployment requires aggressive compression of both vision and language modules. Motivated by these observations, our work specifically targets low-bit quantization, down to 2 bits, for the entire VLM architecture. 

Quantization-aware training (QAT) has been widely demonstrated to preserve accuracy more effectively compared to post-training quantization (PTQ)~\cite{efficientqat,bitdistiller}. However, directly applying QAT methods designed for LLMs to VLMs results in significant accuracy degradation, especially in the case of low-bit quantization. \textbf{Firstly, there is a large discrepancy in training sensitivity between the language module and the vision part.} As illustrated in Figure \ref{fig:grad}, the average absolute gradient value of the LLM is significantly higher than that of the ViT, suggesting a substantial sensitivity discrepancy during training. Treating these modules equally could result in considerable accuracy loss. \textbf{Secondly, QAT demands large calibration sets, yet multimodal data is far more difficult to collect than text-only data.} Alleviating data dependence appears to be crucial for improving the practicality and generalization of VLM quantization. \textbf{Lastly, utilizing QAT on relatively small-scale VLMs in the case of low bit is prone to instability, and often leads to oscillation or poor convergence.} A more robust and targeted optimization strategy is essential to maintain accuracy.

To address the above challenging issues, we built a staged processing with enhanced distillation framework (\textbf{SPEED-Q}) for efficient on-device VLMs quantization. Firstly, we propose a simple yet effective staged quantization strategy to gradually quantize different components, alleviating the sensitivity difference between ViT and LLM. Furthermore, we introduce a novel self-distillation strategy enhanced by an asymmetric clipping method, which improves the initial quantization state, along with a multi-loss optimization approach that enables a more constrained training process. To the best of our knowledge, SPEED-Q is the first quantization framework to enable 2/4-bit weight-only quantization of both vision and language modules in small-scale billion-parameter VLMs, achieving superior accuracy–efficiency trade-offs. As shown in Figure \ref{fig:compare}, our quantized model outperforms existing on-device VLMs by achieving higher accuracy with a smaller model size. For example, 2-bit quantized InternVL3-1B consumes less than 400 MB of running memory while achieving accuracy comparable to the best 0.6B model of FastVLM that requires almost 1.5GB of memory. This advantage unlocks the potential for deploying advanced VLMs on a wider range of edge devices. The main contributions are summarized as follows:
\begin{itemize}
\item For the first time, a 2/4-bit QAT framework is proposed to quantize the entire VLMs with 1-2B parameters, which is essential for on-device deployment.
\item A staged quantization approach is proposed to reduce the effects of heterogeneous training sensitivities across modules in the VLMs.
\item A distillation-enhanced quantization strategy is proposed to reduce data dependence and ultimately mitigate the accuracy collapse commonly observed in low-bit quantization of small VLMs, by employing improved initialization and multiple optimization targets.
\item Extensive evaluations demonstrate that the proposed method consistently surpasses SOTA methods on VLMs of different families, sizes, and quantization schemes.
\end{itemize}

\section{Related Work}

\subsubsection{Mainstream Quantization Paradigms for VLMs.}
Most quantization approaches for VLMs fall into the category of PTQ, exemplified by Q-VLM \cite{qvlm}, MBQ \cite{mbq}, P4Q \cite{p4q}, GPTQ \cite{gptq}, and AWQ \cite{awq}. The PTQ methods generally require minimal calibration data and are computationally efficient. However, significant accuracy degradation occurs when applied to relatively small VLMs. This collapse usually renders those models nonfunctional. QAT integrates quantization in the training loop and has been widely utilized in LLMs. Its extension to VLMs remains limited. The only notable VLM-QAT method, QSLAW \cite{qslaw}, employs group-wise scaling and multimodal warm-up, but only quantizes the language component to 4-bit and leaves the vision module at its original precision (i.e., FP16). Thus, a robust approach enabling accurate low-bit quantization of both vision and language modules in compact VLMs remains an open issue.

\subsubsection{Quantization Sensitivity across Different Modalities.} 
Recent works have shown that vision and language modules in VLMs are unequally susceptible to quantization, prompting several targeted strategies. MQuant \cite{yu2025mquant} employs modality-specific quantization factors, whereas AKVQ-VL \cite{su2025akvq} dynamically adjusts bit budgets within attention to accommodate token-level variation and context saliency. Other component-aware efforts \cite{mbq, hao2024quantized} focus on balancing vision and text losses or addressing specific bottlenecks such as quantized prompts or cache outliers. By focusing solely on cross-modality token effects within the LLM, these approaches overlook the sensitivity of the vision component, limiting their effectiveness in full VLM quantization.

\section{Preliminaries}
Quantization methods for large models often employ group-wise or block-wise quantization, which divides weights into contiguous groups and each group shares a small set of parameters. As observed in \textit{llama.cpp}~\cite{llama_cpp}, further quantizing the group-wise quantization parameters (i.e., scales and zero-points) can yield a strictly lower overall memory footprint with minimal loss of fidelity. We formalize this strategy as \textit{bilevel quantization}.

\subsection{Group-wise Quantization}  
Given a weight tensor $W \in \mathbb{R}^{d_\text{out} \times d_\text{in}}$, we split it into $N$ groups of size $G$, such that $W = [W_1, \ldots, W_N]$, where $W_i \in \mathbb{R}^G$. Each block is quantized independently according to (possibly asymmetric) linear quantization:
\begin{equation}
    \label{eq:quant}
    q_i = \mathrm{clamp}\left(\left\lfloor \frac{w_i}{s} + z \right\rfloor, 0, 2^b-1\right),
\end{equation}
where $s$ is the scale, $z$ is the zero-point, and $b$ is the number of bits. For each quantization block, the scale is defined as $s = (x_\text{max} - x_\text{min})/(2^b - 1)$, mapping the original float range into the range of the targeting bit. The zero point is set as $z = \operatorname{round}(-x_\text{min}/s)$, which preserves the alignment between the original values and their quantized counterparts.

\subsection{Bilevel Quantization}
In practice, storing full-precision quantization parameters for each group requires additional memory, leading to higher resource overhead, especially in low-bit quantization, where a smaller group size is utilized for accuracy maintenance. Bilevel quantization addresses this issue by applying additional group-wise quantization to the first-level quantization parameters.
Specifically, for a weight tensor split into $N$ blocks, we denote the first-level scale vector as $\textbf{S} = [s_1, \ldots, s_N]$ and zero point vector as $\textbf{Z} = [z_1, \ldots, z_N]$ respectively. We then apply a second-level group quantization (with group size $G_q$) to $\textbf{S}$, as follows:
\begin{equation}
\begin{array}{c}
\hat{s} = \mathrm{GroupQuantization}(\textbf{S}, b_s),
\end{array}
\label{eq:sz}
\end{equation}
where $b_s$ is the bit-width for the quantized scales at the second level. Note that zero-points $z_i$ are already rounded to integers during the first quantization stage, and thus are not subjected to further quantization.

\section{Method}

Quantizing VLMs to low bit-widths exposes substantial sensitivity differences between vision and language modules, making unified strategies unstable. We therefore formalize the VLM quantization as minimizing
\begin{equation}
L = f(Q_{\mathrm{ViT}},\ Q_{\mathrm{LLM}};\ S_{\mathrm{stage}},\ S_{\mathrm{opt}}),
\label{eq:total}
\end{equation}
where $Q$ denotes the quantization strategy for each module, $S_{\mathrm{stage}}$ represents the quantization stage arrangement, and $S_{\mathrm{opt}}$ defines the design of optimization objectives. Effective low-bit VLM quantization therefore hinges on module-wise quantization configuration and staged training, paired with robust and data-efficient objectives, to maintain accuracy and training stability.

\subsection{Staged Quantization Strategy}

\begin{figure}[t]
\centering
\includegraphics[width=0.95\columnwidth]{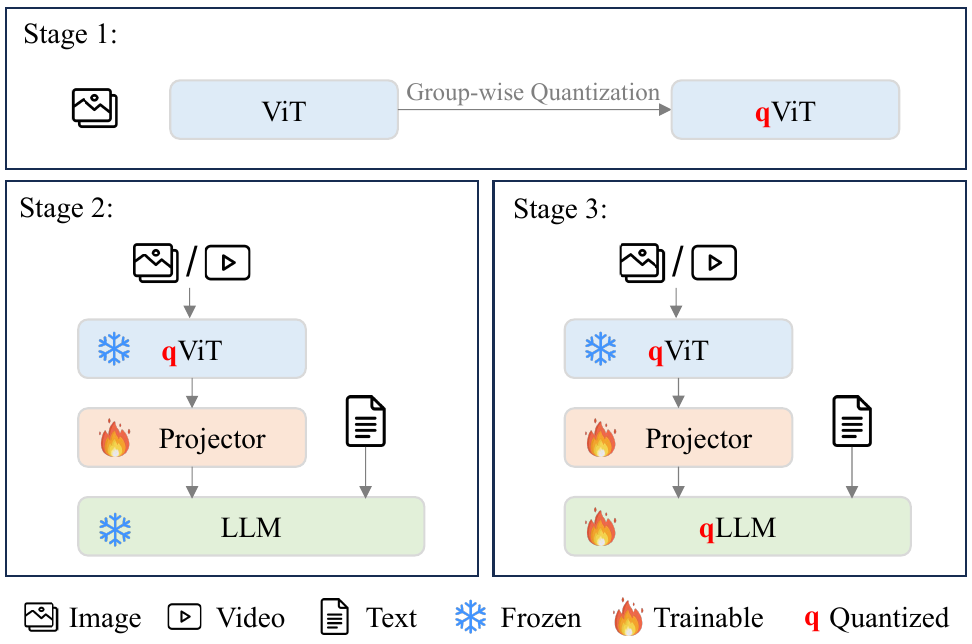}
\caption{Pipeline of the staged quantization strategy. (a) Stage 1: The less sensitive ViT is quantized using an image-only calibration set. (b) Stage 2: Only the projector is trained to better align the quantized ViT (qViT) and the LLM. (c) Stage 3: qViT is frozen, and both the projector and the more sensitive LLM undergo quantization-aware training.}
\label{fig:stage}
\end{figure}

A naive approach for quantizing VLMs is to apply QAT uniformly across all modules. However, our empirical analysis reveals pronounced discrepancies in quantization sensitivity between the vision (ViT) and language (LLM) components: as shown in Figure~\ref{fig:grad}, the language module consistently exhibits higher gradient magnitudes during training, indicating greater vulnerability to low-bit quantization. Motivated by this observation, we design a staged quantization strategy that schedules quantization order adaptively across modules, in order to progressively minimize the end-to-end quantization errors.

Specifically, the staged quantization strategy is illustrated in Figure~\ref{fig:stage}, which decomposes the quantization process into three sequential stages. 
In Stage 1, group-wise quantization strategy is applied to the ViT blocks, where an adaptive rounding mechanism \cite{adaround} is utilized to minimize local reconstruction error for each block.
In Stage 2, we freeze the quantized ViT and train only the Projector to adapt its output, aligning the quantized ViT features with the original distribution expected by the LLM.
In Stage 3, the Projector and quantized LLM are jointly fine-tuned to restore performance and optimize end-to-end alignment under low-bit constraints.

\subsection{Distillation-enhanced Quantization}
\begin{figure}[t]
\centering
\includegraphics[width=0.98\columnwidth]{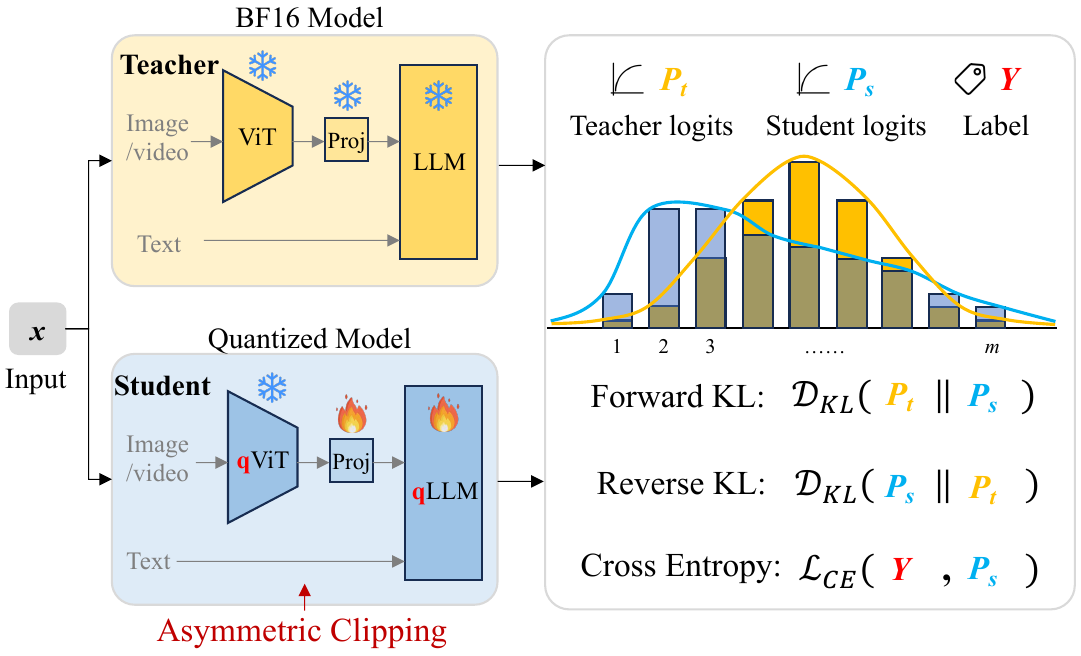}
\caption{Illustration of the distillation-enhanced quantization. Asymmetric clipping provides a more favorable initialization for quantization-aware training. Our objective combines forward KL, reverse KL, and a task-specific loss with ground-truth labels. The self-distillation framework, together with this multi-loss optimization, improves training stability and reduces dependence on large calibration datasets.
}
\label{fig:distill}
\end{figure}

For the quantization of the more sensitive language module, we adopt a distillation-enhanced training procedure that systematically addresses key challenges, including effective initialization, data efficiency, and optimization stability. As shown in Figure \ref{fig:distill}, favorable initial weight distributions for quantization are obtained through asymmetric clipping. We then introduce a self-distillation regime, where the original model guides the training of the quantized model, thus significantly reducing dependence on diverse and large-scale training data. Finally, we combine this with a multi-loss optimization strategy, jointly enforcing consistency with both teacher outputs and ground-truth labels, which further improves convergence and robustness for aggressive low-bit quantization.

\subsubsection{Asymmetric Clipping.}

We perform offline, layer-wise asymmetric clipping as an initialization step before QAT, aiming to stabilize training while incurring minimal computational overhead. Specifically, given input $x$ from a small calibration set, we automatically search for two optimal clipping values $\alpha$ and $\beta$ for each layer of the model. These values aim to minimize the $L_2$ distance between the outputs of quantized and original models:

\begin{equation}
    \label{eq:ab}
    \alpha^*,\beta^* = \underset{\alpha,\beta}{\mathrm{arg\,min} } \,\,\,\, \left \|  \widetilde{\mathrm{W}_c }x -\mathrm{W}x \right \|,
\end{equation}
where $\mathrm{W}_c=clip(\mathrm{W},\alpha,\beta)$ clips the weights out of the range $[\alpha, \beta]$, and $\widetilde{\mathrm{W}_c }$ denotes the quantized weights. In this way, the impact of outliers in the initial floating-point weights can be mitigated to some extent.

\subsubsection{Self-distillation Strategy.}
As shown in Figrue \ref{fig:distill}, 
we employ self-distillation, where the fixed pre-trained BF16 model serves as the teacher and the quantized model as the student. 
The distillation loss combines the reverse KL and the forward KL as follows:
\begin{equation}
    \label{eq:kl}
    \mathcal{L}_{distill}=\gamma\mathcal{D}_{KL}(P_s || P_t)+(1-\gamma)\mathcal{D}_{KL}(P_t || P_s),
\end{equation}
where $P_t$ and $P_s$ represent the output logits of the teacher model and the student model, respectively. The coefficient $\gamma$ is estimated by the averaged token probability on a randomly selected mini-batch of training samples during the early stage of training, formalized as:
\begin{equation}
    \label{eq:gama}
    \gamma =\mathbb{E}_{(x,y)\sim{\mathbb{D} }} [\frac{1}{\left | \left \{  y\right \}  \right | }\sum_{i=1}^{\left | \left \{  y\right \}  \right |}P_t(y_i|x,y_{<i} )].
\end{equation}
$\gamma$ serves as a weighting factor that balances the contributions of forward and reverse KL divergences in the distillation loss. 
When the teacher is highly confident, reverse KL encourages the student to concentrate on strong predictions; otherwise, forward KL is prioritized to better match the full distribution.
This adaptive mechanism leads to more stable and effective knowledge transfer, especially in low-bit quantization scenarios.

\subsubsection{Multi-loss Optimization Strategy.}
We extend our self-distillation framework by incorporating explicit supervision from ground-truth labels to directly regularize the model’s output distribution. This dual objective anchors the optimization process to the target task, ensuring that the quantized model remains aligned with the original learning target. We introduce the standard cross-entropy loss for the target task, and the total loss is then defined as
\begin{equation}
    \label{eq:loss2}
    \mathcal{L}_{total}=\lambda \mathcal{L}_{distill}+\zeta \mathcal{L}_{CE}(Y, P_s).
\end{equation}

These two losses are complementary: the former stabilizes feature-space alignment via self-distillation and reduces reliance on large-scale datasets, while the latter ensures fidelity to the target task. Together, they jointly guide the optimization trajectory, significantly alleviating training oscillations and instability. Although alternative weighting schemes of the two losses may yield marginally better performance, we set $\lambda = \zeta = 1$ for simplicity.

\begin{table*}[ht]
  \centering
  \small
  \setlength{\tabcolsep}{0.6mm} 
      \begin{tabular}{cclccccccccc}
        \toprule
        Model & Bitwidth & \multicolumn{1}{c}{Method} & MMBench	& MMStar & MMMU &	Hallusion.	& AI2D	& OCRBench & SEED & ScienceQA  & Avg. ($\uparrow$) \\
        \midrule
        \multirow{10}{*}{InternVL3-1B} & BF16 & - & 69.08 &	52.27 &	41.33 &	36.25 &	69.75 &	79.7 &	71.16 &	89.84 &	63.67 \\
        \cmidrule(lr){2-12}
        ~ & \multirow{5}{*}{4-bit} & RTN & 63.16 &	48.73 &	36.67 &	32.89 &	65.12 &	74.1 &	69.01 &	83.21 &	59.11 \\
        ~ & ~ & GPTQ (ICLR'23) & 62.77 &	\textbf{52.2} &	35 &	32.69 &	63.73 &	75.6 &	68.99 &	83.98 &	59.37 \\
        ~ & ~ & AWQ (MLSys'24) & 62.62 &	49.8 &	36.56 &	35.95 &	64.67 &	74.8 &	68.59 &	83.36 &	59.54 \\
        ~ & ~ & MBQ (CVPR'25) & 63.85 &	50.93 &	38.67 &	31.66 &	66.87 &	74.9 &	69.28 &	82.88 &	59.88 \\
        ~ & ~ & \textbf{SPEED-Q (Ours)$\ddagger$} & \textbf{66.68} &	51.07 &	\textbf{40.44} & \textbf{37.01} &	\textbf{68.81} & \textbf{79.7} & \textbf{71.39} & \textbf{84.41} &	\textbf{62.44} \\
        \cmidrule(lr){2-12}
        ~ & \multirow{4}{*}{2-bit} & RTN & 0.04 & 7.6 & 6.89 & 0.73 & 6.67 & 0.5 & 6.45 & 8.01 & 4.61 \\
        ~ & ~ & GPTQ (ICLR'23) & 0.15 & 10.2 & 11.67 & 0.43 & 8.19 & 1.6 & 11.12 & 12.83 & 7.02 \\
        ~ & ~ & MBQ (CVPR'25) & 0.00 &	7.53 &	5.33 &	0.11 &	5.25 &	0.1 &	6.30 &	5.87 &	3.81 \\
        ~ & ~ & \textbf{SPEED-Q (Ours)$\ddagger$} & \textbf{53.13} &	\textbf{43.4} &	\textbf{32.89} &	\textbf{40.21} &	\textbf{57.58} &	\textbf{58.5} &	\textbf{66.93} &	\textbf{59.18} &	\textbf{51.48} \\
    
        \midrule

        \multirow{10}{*}{InternVL3-2B} & BF16 & - & 78.68 &	61.07 &	45.56 &	41.94 &	78.76 &	83.6 &	74.95 &	95.23 &	69.97 \\
        \cmidrule(lr){2-12}
        ~ & \multirow{5}{*}{4-bit} & RTN & 73.57 &	57 &	43.33 &	39.37 &	75.91 &	81.2 &	73.22 &	92.32 &	66.99 \\
        ~ & ~ & GPTQ (ICLR'23) & 75.93 &	57.93 &	45.89 &	42.72 &	75.94 &	81.2 &	73.89 &	92.42 &	68.24 \\
        ~ & ~ & AWQ (MLSys'24) & 75.12 &	58.13 &	\textbf{47.22} &	41.50 &	76.88 &	82.1 &	73.85 &	91.23 &	68.25 \\
        ~ & ~ & MBQ (CVPR'25) & 75.43 &	\textbf{59.47} &	46.11 &	38.25 &	76.36 &	82.0 &	73.37 &	92.75 &	67.97 \\
        ~ & ~ & \textbf{SPEED-Q (Ours)$\ddagger$} & \textbf{77.67} &	58.47 &	44.56 &	\textbf{44.03} &	\textbf{77.3} &	\textbf{82.4} &	\textbf{74.42} &	\textbf{93.32} &	\textbf{69.02} \\
        \cmidrule(lr){2-12}
        ~ & \multirow{4}{*}{2-bit} & RTN & 0.08 & 2.2 & 1.89 & 1.24 & 1.49 & 0.3 & 6.89 & 1.81 & 1.99 \\
        ~ & ~ & GPTQ (ICLR'23) & 0.62 & 10.53 & 7.11 & 3.20 & 11.43 & 12.7 & 12.63 & 12.4 & 8.83 \\
        ~ & ~ & MBQ (CVPR'25) & 0.12 & 9.47 & 8.44 & 3.48 & 9.29 & 5.0 & 10.18 & 10.35 & 7.04 \\
        ~ & ~ & \textbf{SPEED-Q (Ours)$\ddagger$} & \textbf{68.30} &	\textbf{49.13} &	\textbf{40.56} &	\textbf{38.34} &	\textbf{69.95} &	\textbf{64.9} &	\textbf{72.34} &	\textbf{72.20} &	\textbf{59.47} \\
    
        \midrule
        
        \multirow{10}{*}{Qwen2-VL-2B} & BF16 & - & 71.59 &	46.4 &	39.44 &	41.88 &	72.02 &	81.0 &	72.66 &	73.68 &	62.33 \\
        \cmidrule(lr){2-12}
        ~ & \multirow{5}{*}{4-bit} & RTN & \textbf{70.98} &	45.07 &	37.22 &	\textbf{42.74} &	\textbf{71.15} &	78.8 &	72.45 &	70.67 &	61.13 \\
        ~ & ~ & GPTQ (ICLR'23) & 70.51 &	46.33 &	36.22 &	39.62 &	70.53 &	79.5 &	72.62 &	72.10 &	60.93 \\
        ~ & ~ & AWQ (MLSys'24) & 68.89 &	44.8 &	37.33 &	39.55 &	70.08 &	78.9 &	71.83 &	\textbf{72.15} &	60.44 \\
        ~ & ~ & MBQ (CVPR'25) & 70.55 &	44.53 &	38.22 &	39.89 &	70.21 &	\textbf{80.9} &	71.86 &	71.72 &	60.98 \\
        ~ & ~ & \textbf{SPEED-Q (Ours)$\ddagger$} & 69.85 &	\textbf{50.87} &	\textbf{42.0} &	41.71 &	70.92 &	76.5 &	\textbf{74.40} &	72.01 &	\textbf{62.28} \\
        \cmidrule(lr){2-12}
        ~ & \multirow{4}{*}{2-bit} & RTN & 1.16 & 15.6 & 12.0 & 16.11 & 18.91 & 10.9 & 17.72 & 24.65 & 14.63 \\
        ~ & ~ & GPTQ (ICLR'23) & 1.97 & 17.33 & 13.33 & 3.97 & 16.09 & 20.1 & 23.19 & 24.51 & 15.06 \\
        ~ & ~ & MBQ (CVPR'25) & 2.17 & 17.53 & 13.56 & 26.30 & 18.65 & 20.4 & 23.46 & 26.13 &  18.52
        \\
        ~ & ~ & \textbf{SPEED-Q (Ours)$\ddagger$} & \textbf{57.31} &	\textbf{42.93} &	\textbf{35.22} &	\textbf{34.11} &	\textbf{60.33} &	\textbf{57.0} &	\textbf{68.51} &	\textbf{60.13} &	\textbf{51.94} \\
        \bottomrule
      \end{tabular}
  \caption{Main results on InternVL3 and Qwen2-VL families. $\ddagger$ indicates that both ViT and LLM are quantized; otherwise, only the LLM is quantized. To ensure a fair comparison, competing methods use 4-bit weight quantization with group size 128 (w4g128), while SPEED-Q employs \textit{bilevel quantization} with w4g32g128, yielding a comparable average bitwidth (4.25 vs. 4.28 bits) calculated on the quantized components in VLMs. Under 2-bit settings, the compared methods exhibit severe accuracy degradation across various configurations. We thus report them under w2g16, while SPEED-Q uses w2g16g16.}
  \label{tab:sota}
\end{table*}

\begin{table}[t]
    \centering
    \small
    \setlength{\tabcolsep}{1.2mm} 
        \begin{tabular}{clccc}
            \toprule
            Model & Method & BF16 & 4-bit & 2-bit \\
            \midrule  
            \multirow{2}{*}{InternVL3-1B} & Quantize LLM only & \multirow{2}{*}{1789.5} & 941.7 & 853.1 \\
            ~ & \textbf{SPEED-Q(Ours)} & ~ & \textbf{485.0} & \textbf{315.0}  \\

            \midrule  
            \multirow{2}{*}{InternVL3-2B} & Quantize LLM only & \multirow{2}{*}{3984.5} & 1454.6 & 1218.7 \\
            ~ & \textbf{SPEED-Q(Ours)} & ~ &\textbf{936.1} & \textbf{661.5} \\
            \bottomrule
        \end{tabular}
    \caption{Model file sizes (in MB) under different quantization schemes. SPEED-Q quantizes both ViT and LLM, enabling smaller model file sizes.}
    \label{tab:filesize}
\end{table}

\begin{table}[t]
    \centering
    \small
    \setlength{\tabcolsep}{0.8mm} 
        \begin{tabular}{cclcc}
            \toprule
             Model & Bitwidth & \multicolumn{1}{c}{Method} & DocVQA & RealWorldQA \\
            \midrule
            \multirow{3}{*}{LLaVA-13B} & BF16 & - & 14.46 & 48.49 \\
            \cmidrule(lr){2-5}
             & \multirow{2}{*}{4-bit} & QSLAW  & 3.46 & 40.65 \\
             &  & \textbf{SPEED-Q$\ddagger$} & \textbf{13.54} & \textbf{55.16} \\
            \bottomrule
        \end{tabular}
    \caption{Comparison with QSLAW \cite{qslaw}. Since QSLAW quantizes LLaVA-13B \cite{llava} only to 4-bit, we perform comparisons under the same bitwidth setting. $\ddagger$ indicates that both ViT and LLM are quantized.}
    \label{tab:qslaw}
\end{table}

\section{Experiments}

\begin{table*}[ht]
  \centering
  \small
  \setlength{\tabcolsep}{0.5mm} 
      \begin{tabular}{lcccccccccc}
        \toprule
        Model & Bitwidth & MMBench	& MMStar & MMMU &	Hallusion.	& AI2D	& OCRBench & SEED & ScienceQA  & Avg. ($\uparrow$) \\
        \midrule
        \textit{BF16:} & ~ \\
        SmolVLM-256M~\cite{smolvlm} &	BF16 & 25.19 &	34.6 & 27.0 & 26.33 & 47.09 & 52.6 & 54.31 & 72.39 &	42.44  \\
        SmolVLM-500M~\cite{smolvlm} &	BF16 & 41.06 & 38.33 & 31.44 & 29.43 & 59.52 & 61.0 & 62.03 & 76.73 & 49.94  \\
        FastVLM-600M~\cite{fastvlm} &	BF16 & 56.39 & 44.8 & 33.55 & 37.11 & 67.91 & 58.2 & 57.69 & 80.54 & \textbf{54.52} \\
        H2OVL-800M~\cite{h2ovl} & BF16 & 48.14 & 38.93 & 30.78 & 28.60 & 53.47 & 75.0 & 60.18 & 66.48 & 50.20  \\
        
        \midrule
        
        \textit{Quantized:} & ~ \\
        \textbf{InternVL2.5-1B SPEED-Q} &	\textbf{4-bit}  & \textbf{66.41} & \textbf{50.27} & \textbf{36.89} & \textbf{41.41} & \textbf{68.36} & \textbf{75.8} & \textbf{71.47} & \textbf{89.08} & \textbf{62.46}   \\
        InternVL2.5-1B SPEED-Q &	2-bit & 53.68	& 43.6	& 33.33	& 37.04	& 59.42	& 61.2	& 66.78 & 61.18	& 52.03  \\
        FastVLM-600M SPEED-Q &	4-bit  & 51.70	& 42.4 & 32.22 & 31.71 & 64.70 & 53.4 & 61.43 & 77.68 & 51.91   \\
        FastVLM-600M SPEED-Q &	2-bit  & 41.99	& 39.53	& 29.56	& 31.93	& 54.86	& 50.2	& 60.72	& 61.66	& 46.31   \\
        \bottomrule
      \end{tabular}
  \caption{Main results on on-device VLMs. 4-bit InternVL2.5-1B (485 MB, model file size) outperforms BF16 FastVLM (1517 MB) by +7.94 points, while the 2-bit InternVL2.5-1B (315 MB) matches its performance at 1/5 size.}
  \label{tab:on-device}
\end{table*}

\begin{table}[t]
    \centering
    \small
    \setlength{\tabcolsep}{1mm} 
        \begin{tabular}{ccccc}
            \toprule
            \multicolumn{2}{c}{Components} & \multirow{2}{*}{MMStar} & \multirow{2}{*}{AI2D} & \multirow{2}{*}{OCRBench}\\
            \cmidrule(lr){1-2}
            Task Loss & Distillation Loss & ~ & ~ & ~ \\
            \midrule            
            \ding{51} & \ding{55} & 49.33 & 66.13 & 75.5 \\
            \ding{55} & \ding{51} & 50.40 & 68.46 & 77.9\\
            \ding{51} & \ding{51} & \textbf{51.07} & \textbf{68.81} & \textbf{79.7} \\
            \midrule
            \multicolumn{2}{c}{Joint Quantization} & 49.67 & 68.46 & 77.0 \\
            \multicolumn{2}{c}{Staged Quantization} & \textbf{51.07} & \textbf{68.81} & \textbf{79.7} \\ 
            \bottomrule
        \end{tabular}
    \caption{Ablation study on various configurations of training loss and training strategy in our approach.}
    \label{tab:ablation}
\end{table}

\subsection{Experimental Setups}
\subsubsection{Implementation Details.}
SPEED-Q is implemented using DeepSpeed \cite{deepspeed} and applies quantization to all linear layers in both the ViT and LLM. All training data used in SPEED-Q are drawn from publicly available open-source datasets. The detailed description is included in the supplementary materials. During quantization-aware training, we sample 10\% of the data from each dataset, resulting in approximately 680,000 training samples in total. The quantization scheme follows the \textit{bilevel quantization} described in Section~3. For 4-bit quantization, the first-level group size is 32 and the second-level group size is 128. For 2-bit quantization, both levels use a group size of 16.

\subsubsection{Evaluation Datasets.}
To evaluate the performance of the quantized model, we conduct experiments on various benchmarks based on the VLMEvalKit~\cite{vlmevalkit}. Specifically, we use MMBench~\cite{mmbench} and MMStar~\cite{mmstar} for comprehensive multimodal evaluation, ScienceQA~\cite{scienceqa} and MMMU~\cite{mmmu} to evaluate visual reasoning, AI2D~\cite{ai2d} and OCRBench~\cite{ocrbench} for text recognition and comprehension, SEED-Bench~\cite{seed-bench} to test visual perception, and HallusionBench~\cite{hallusionbench} for hallucination evaluation.

\subsection{Comparison with State-of-the-Arts}

Table~\ref{tab:sota} presents a performance comparison between our SPEED-Q and previous state-of-the-art methods across multiple VLM families and benchmarks. Generally speaking, our method quantizes more components while delivering superior performance. Specifically, by utilizing the proposed SPEED-Q, the quantized 4-bit model exhibits a marginal performance decrease of 3\% compared to their original BF16 counterpart. This observation is consistent across InternVL3-1B, InternVL3-2B, and Qwen2-VL-2B. Compared to other methods that only quantize the LLM module, SPEED-Q achieves superior performance on most of the evaluated benchmarks. Furthermore, nearly all existing methods suffer severe accuracy degradation under 2-bit quantization, rendering them practically ineffective. In contrast, SPEED-Q maintains controllable performance degradation under 2-bit quantization, with accuracy dropping by only approximately 15\% with respect to the BF16 models. This result further demonstrates the effectiveness and robustness of our method in extreme low-bit. Since the visual features extracted by the ViT encoder are crucial for multimodal understanding, the results indicate that our method effectively mitigates the impact of ViT quantization on the end-to-end performance. 

Table~\ref{tab:filesize} lists the model size of different approaches, which is critical for memory usage and inference speed when deployed on edge devices. SPEED-Q successfully compresses VLMs into significantly smaller model sizes by simultaneously quantizing ViT and LLM components. 

Table~\ref{tab:qslaw} compares the QAT-based method QSLAW with our method on DocVQA \cite{docvqa} and RealWorldQA \cite{realworldqa}. As shown in the table, SPEED-Q demonstrates clear superiority.

\begin{table*}[ht]
  \centering
  \small
  \setlength{\tabcolsep}{0.6mm} 
      \begin{tabular}{cclccccccccc}
        \toprule
        Model & Bitwidth & \multicolumn{1}{c}{Method} & MMBench	& MMStar & MMMU &	Hallusion.	& AI2D	& OCRBench & SEED & ScienceQA  & Avg. ($\uparrow$) \\
        \midrule
        \multirow{5}{*}{InternVL3-8B} & BF16 & - & 85.22 & 68.6 & 56.89 & 48.24 & 85.36 & 88.2 & 77.28 & 97.85 & 75.95 \\
        \cmidrule(lr){2-12}
        ~ & \multirow{2}{*}{4-bit} & MBQ (CVPR'25) & \textbf{84.75}	& \textbf{67.4}	& \textbf{55.67}	& 48.92	& 84.10	& \textbf{87.4}	& \textbf{77.09}	& \textbf{97.47}	& \textbf{75.35} \\
        ~ & ~ & \textbf{SPEED-Q (Ours)$\ddagger$} & 84.25	& 66.47	& 54.78	& \textbf{51.99}	& 84.10	& 84.6	& 76.98	& 97.19	& 75.04 \\
        \cmidrule(lr){2-12}
        ~ & \multirow{2}{*}{2-bit} & MBQ (CVPR'25) & 66.29	& 53.13	& 43.44	& 38.75	& 71.57	& 70.0	& 70.83	& 79.30	& 61.66 \\
        ~ & ~ & \textbf{SPEED-Q (Ours)$\ddagger$} & \textbf{75.81}	& \textbf{60.2}	& \textbf{46.22}	& \textbf{50.61}	& \textbf{79.08}	& \textbf{75.8}	& \textbf{75.24}	& \textbf{84.69}	& \textbf{68.46} \\
        \bottomrule
      \end{tabular}
  \caption{Main results of SPEED-Q and MBQ on InternVL3-8B.}
  \label{tab:gen}
\end{table*}

\subsection{Comparison with Lightweight VLMs}

As shown in Table~\ref{tab:on-device}, we first compare the quantized models with the existing VLMs designed for edge deployment, such as SmolVLM, FastVLM, and H2OVL, which typically have fewer than 1B parameters. The 4-bit quantized InternVL2.5-1B achieves substantial leadership across multiple benchmarks compared to all the existing BF16 lightweight models. Furthermore, the 2-bit quantized InternVL2.5-1B exhibits comparable accuracy to FastVLM, which exhibits a favorable trade-off between model size and performance. More importantly, the quantized model significantly reduces computational resource requirements. In addition, we apply SPEED-Q on the FastVLM. SPEED-Q successfully limits the performance drop to 4.78\% and 15.06\% under 4-bit and 2-bit quantization respectively. The results further demonstrate that SPEED-Q incurs manageable accuracy degradation, even on the lightweight VLMs.

\begin{figure}[t]
\centering
\includegraphics[width=0.97\columnwidth]{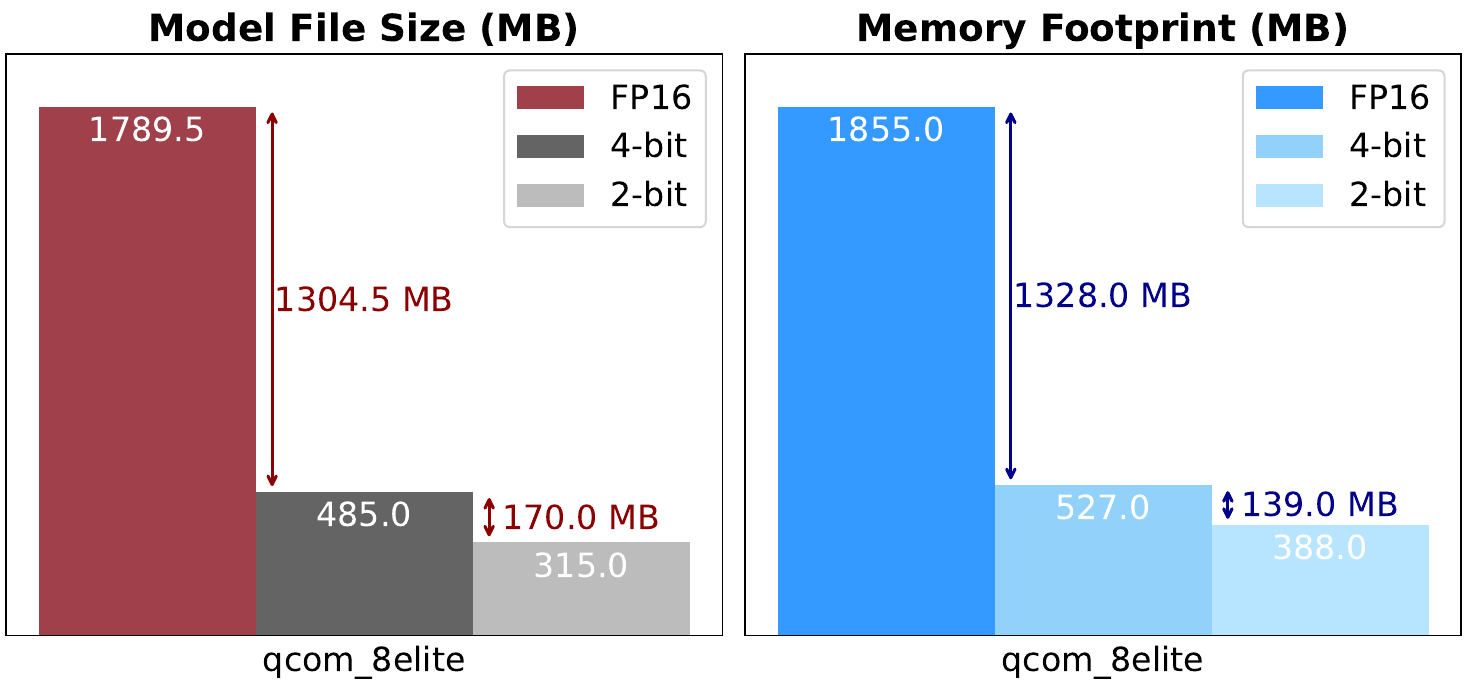}
\caption{On-device efficiency evaluation of FP16 and low-bit InternVL2.5-1B on Samsung Galaxy S25 Ultra.}
\label{fig:efficiency}
\end{figure}

\subsection{Ablation Study}

To evaluate the contribution of each component, we conduct ablation studies on the 4-bit quantization of InternVL3-1B using the MMStar, AI2D, and OCRBench benchmarks.

\subsubsection{Effectiveness of Multi-loss Optimization Strategy.} As shown in Table~\ref{tab:ablation}, using distillation loss results in better accuracy preservation. Jointly optimizing both losses achieves significant performance improvements across all three benchmarks. This result confirms that our multi-loss optimization strategy enhances training stability by identifying more effective quantization targets.

\subsubsection{Effectiveness of the Staged Quantization Strategy.} As shown in Table~\ref{tab:ablation}, simultaneously quantizing both the ViT and LLM (i.e., joint quantization) often leads to unstable training and suboptimal performance. In contrast, staged quantization strategy applies quantization to the ViT and LLM in a sequential, multi-stage way. The results demonstrate that the staged quantization strategy achieves superior accuracy across multiple benchmarks. It confirms our hypothesis that decoupling the quantization process reduces optimization difficulty, improves training stability, and ultimately yields higher-quality quantized models.

\begin{figure*}[t]
\centering
\includegraphics[width=0.97\textwidth]{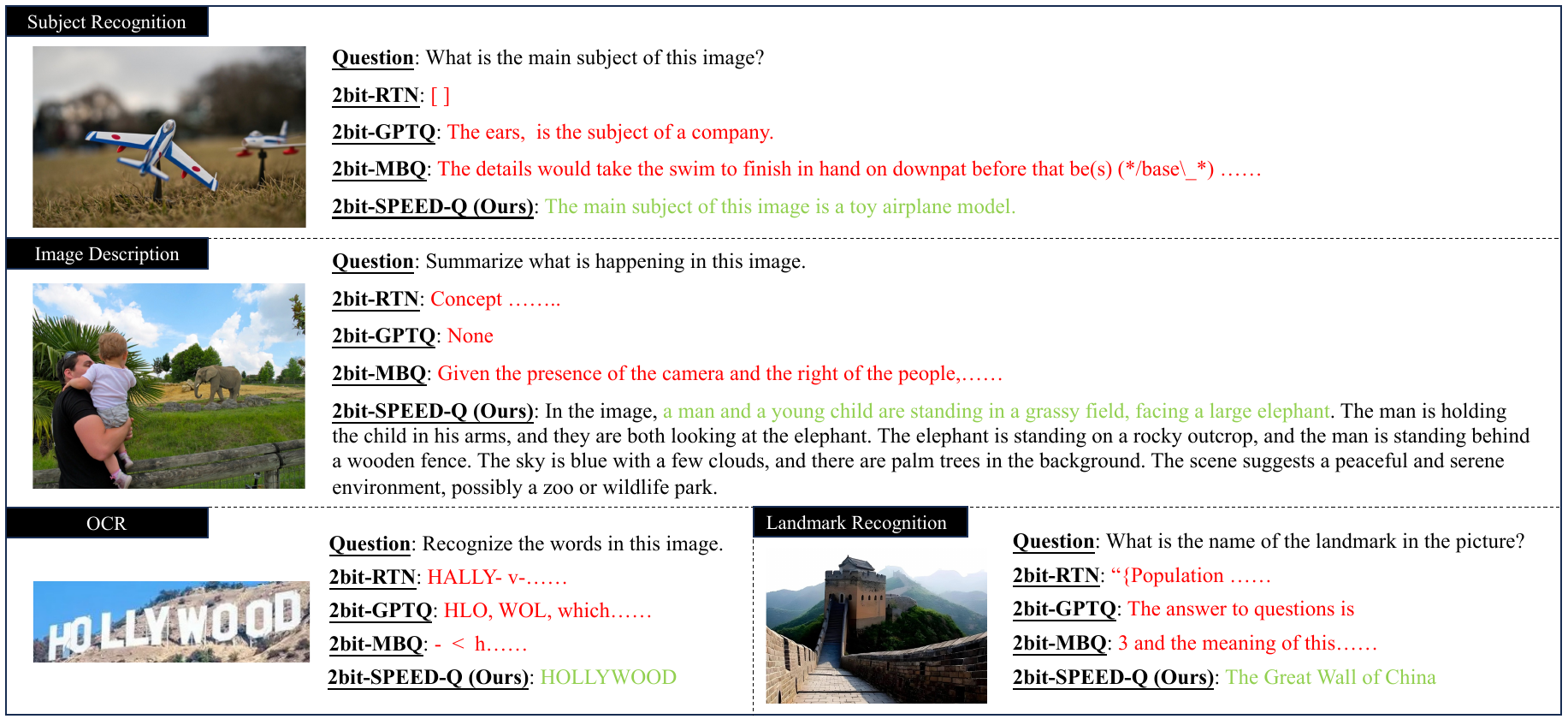}
\caption{Qualitative comparison on InternVL2.5-1B under 2-bit quantization. Our method generates more accurate responses than RTN, GPTQ, and MBQ. Correct (green) and incorrect (red) parts are highlighted.}
\label{fig:case}
\end{figure*}

\subsection{Generalization on Large VLMs}

To further assess the generalization of our method, we evaluate its effectiveness on InternVL3-8B with almost 8 billion parameters. As shown in Table~\ref{tab:gen}, both SPEED-Q and MBQ preserve the accuracy of the original model, with SPEED-Q further distinguishing itself by quantizing both the vision and language components. When the model is quantized to 2-bit, SPEED-Q demonstrates superior performance across multiple benchmarks.

\subsection{Efficiency Evaluation}

Figure~\ref{fig:efficiency} illustrates the benefits of low-bit quantization in reducing both model file size and runtime memory footprint. Compared to the FP16 baseline, 4-bit quantization reduces the model size by 1304.5 MB, with an additional 170.0 MB reduction under 2-bit quantization. This advantage makes the deployment of VLMs more practical by requiring less storage and reducing distribution costs. Moreover, lower-bit quantization further reduces runtime memory footprint, which is crucial for enabling efficient and stable inference on resource-constrained edge devices.

\subsection{Qualitative Analysis}

As shown in Figure~\ref{fig:case}, SPEED-Q generates meaningful responses, while existing methods fail to produce accurate outputs. The observation is consistent across different tasks, demonstrating superior preservation of semantic and visual understanding under extreme low-bit quantization.

\section{Conclusion}
In order to further unleash the potential of VLMs deployment on edge devices, we have presented SPEED-Q, a novel quantization framework for 1-2B parameter VLMs. Our staged quantization strategy addresses the divergent training sensitivities between ViT and LLM modules, enabling stable and effective quantization across heterogeneous modalities. Furthermore, we propose distillation-enhanced quantization to stabilize the training process of low-bit VLMs and reduce dependence on large labeled datasets. Extensive experiments demonstrate that SPEED-Q achieves state-of-the-art performance. These results significantly advance the practical deployment of VLMs on edge devices. In future work, we plan to extend SPEED-Q from weight-only to weight-activation quantization, aiming to achieve further improvements in inference efficiency.

\bibliography{aaai2026}

\newpage
\appendix

\twocolumn[
  \begin{@twocolumnfalse}
  \medskip
  \centering
  \LARGE\bfseries Appendix \par
  \medskip
  \medskip
  \medskip
  \medskip
  \end{@twocolumnfalse}
]

For our SPEED-Q, this appendix provides additional implementation details and experimental results that were omitted from the main paper due to space constraints.

\section{Training Data Details}

We construct the training dataset by curating samples from InternVL~\cite{internvl}, encompassing four modalities: text-only, single-image, multi-image, and video-based inputs, all derived from publicly accessible sources. The resulting corpus spans a wide variety of task categories, supporting robust model generalization across heterogeneous input modalities and downstream tasks. Dataset statistics and composition are detailed in Table~\ref{tab:data}. For quantization-aware training, we randomly sample 10\% of each dataset, yielding a representative \textit{training subset} of $\sim$680K examples with broad coverage across tasks and domains.

\section{More Implementation Details}

All experiments are conducted on 8 NVIDIA A100 GPUs. For InternVL3-1B, Stage 1 performs calibration for ViT quantization using a randomly sampled calibration set of 128 images, taking less than 1 hour. Stage 2 warms up the projector using 10\% of the \textit{training subset} and lasts about 3 hours. Stage 3 conducts quantization-aware training on the \textit{training subset}, which takes roughly 36 hours. The AdamW optimizer is used with zero weight decay and a constant learning rate of $8\times10^{-6}$. The self-distillation balancing factor $\gamma$ is determined based on the average prediction confidence of the floating-point teacher model, computed over the first 100 training steps. A fixed random seed of 42 is used to ensure reproducibility.

\section{Additional Results on InternVL2.5}

Complete quantization results for InternVL2.5-1B are presented in Table~\ref{supp-tab1}. Together with the results in the main paper, SPEED-Q consistently outperforms existing state-of-the-art methods under both 2-bit and 4-bit quantization, demonstrating significant and robust performance gains over previous approaches.

\begin{table}[t]
    \centering
    \small
    \setlength{\tabcolsep}{0.5mm} 
        \begin{tabular}{lll}
            \toprule
            Dataset & Size & Data Class\\
            \midrule
            \textit{Single-Image: } \\
            ALLaVA \cite{allava} &	1037218 &  Conversation  \\
            ShareGPT4o \cite{sharegpt4o} & 57289 &  Captioning  \\
            ShareGPT4V \cite{sharegpt4v} & 767083 &  General QA \\
            LLaVA-Instruct \cite{llava_instruct} & 157712 & General QA \\
            MMInstruct \cite{mminstruct} & 378186 &  General QA \\
            DVQA \cite{dvqa} & 200000 &  Chart \\
            ChartQA \cite{chartqa} & 18317 &  Chart \\
            AI2D \cite{ai2d} & 12413 &  Science \\
            DocVQA \cite{docvqa} & 10211 &  Document \\
            GeoQA+ \cite{geoqa} & 72318 &  Mathematics \\
            SynthDoG-EN \cite{synthdog} & 29765 &  OCR \\
            SROIE \cite{sroie} & 33626 &  OCR \\
            ScreenQA \cite{screenqa} & 209098 & GUI \\

            \midrule
            \textit{Multi-Image: } \\
            Spot-the-Diff \cite{mantis} & 8007 & General QA \\
            MultiVQA \cite{mantis} & 4993 & General QA \\
            Birds-to-Words \cite{mantis} & 2649 & General QA \\
            NLVR2 \cite{nlvr2} & 86373 & General QA \\

            \midrule
            \textit{Video: } \\
            ShareGPT4o-Video \cite{sharegpt4o} & 2111 &  Captioning \\
            LLaVA-Video \cite{llava_video} & 898992 & General QA \\

            \midrule
            \textit{Text: } \\
            Firefly \cite{firefly} & 1649399 & General QA \\
            WizardLM \cite{wizardlm} & 70000 & General QA \\
            Magpie-Qwen2-Pro \cite{magpie} & 1000000 & General QA \\
            Long-Instruction \cite{long} & 9471 & Long Context \\
            MathQA \cite{mathqa} & 35000 & Mathematics \\
            Code-Feedback \cite{code} & 66383 & Code \\

            \midrule
            Total & 6816614   \\
            
            \bottomrule
        \end{tabular}
    \caption{Details of the training data.}
    \label{tab:data}
\end{table}

\section{More Qualitative Results}

\begin{figure*}[h]
\centering
\includegraphics[width=0.95\textwidth]{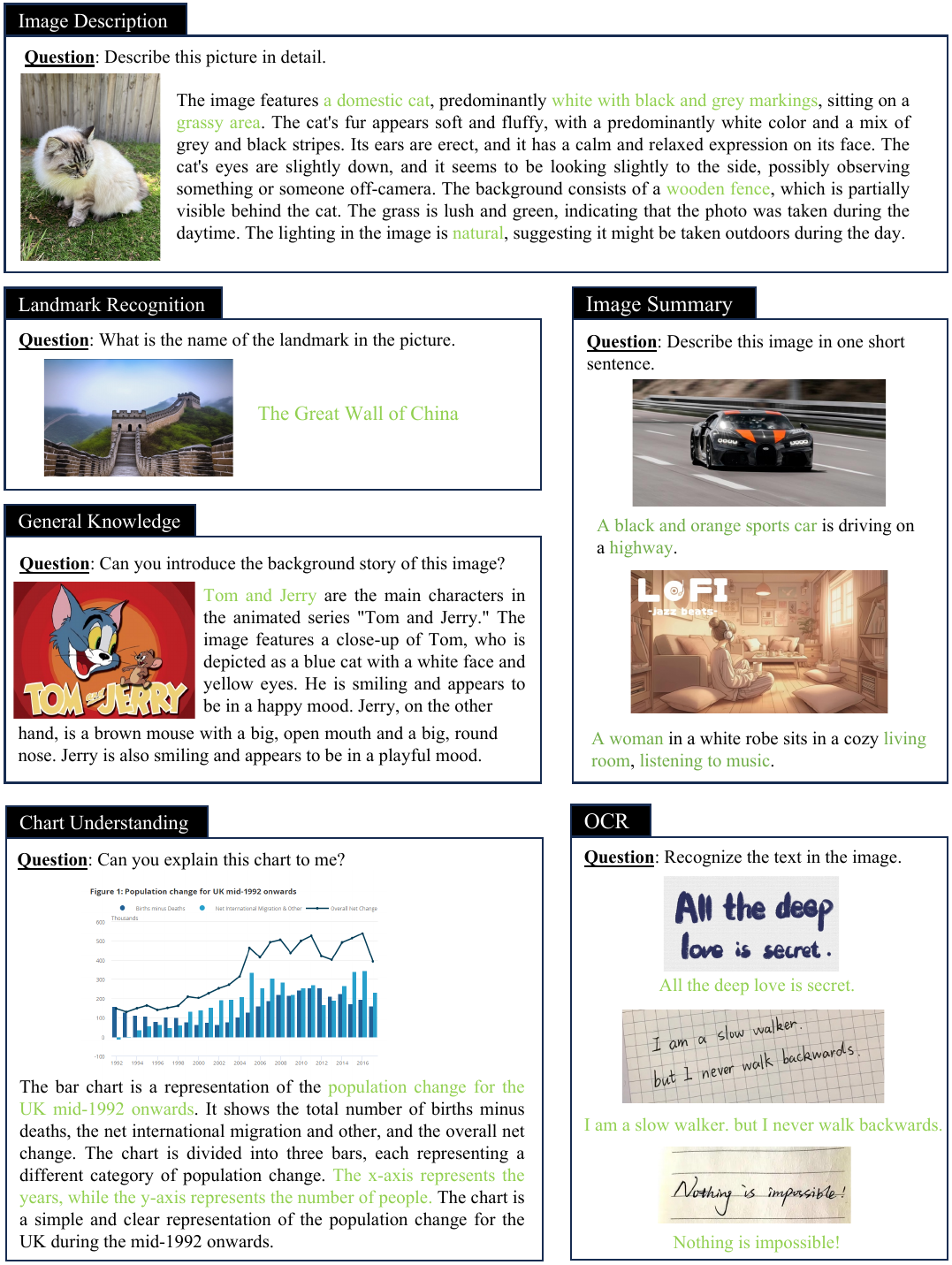}
\caption{General question answering capability of InternVL2.5-SPEED-Q(2-bit). Our model has multi-faceted capabilities, including recognition of landmarks, image summary, answering questions about general knowledge, understanding charts, recognizing text, and more.}
\label{fig:page}
\end{figure*}

We present qualitative examples across a range of vision-language tasks to illustrate the effectiveness of our method under 2-bit quantization. As shown in Figure~\ref{fig:page}, InternVL2.5-SPEED-Q (2-bit) produces accurate and coherent responses on diverse inputs, preserving rich semantic knowledge and demonstrating robust multimodal understanding. These results highlight the capability of SPEED-Q to maintain high-quality generation even in extremely low-bit, confirming its practical viability.

\begin{figure*}[h]
\centering
\includegraphics[width=0.95\textwidth]{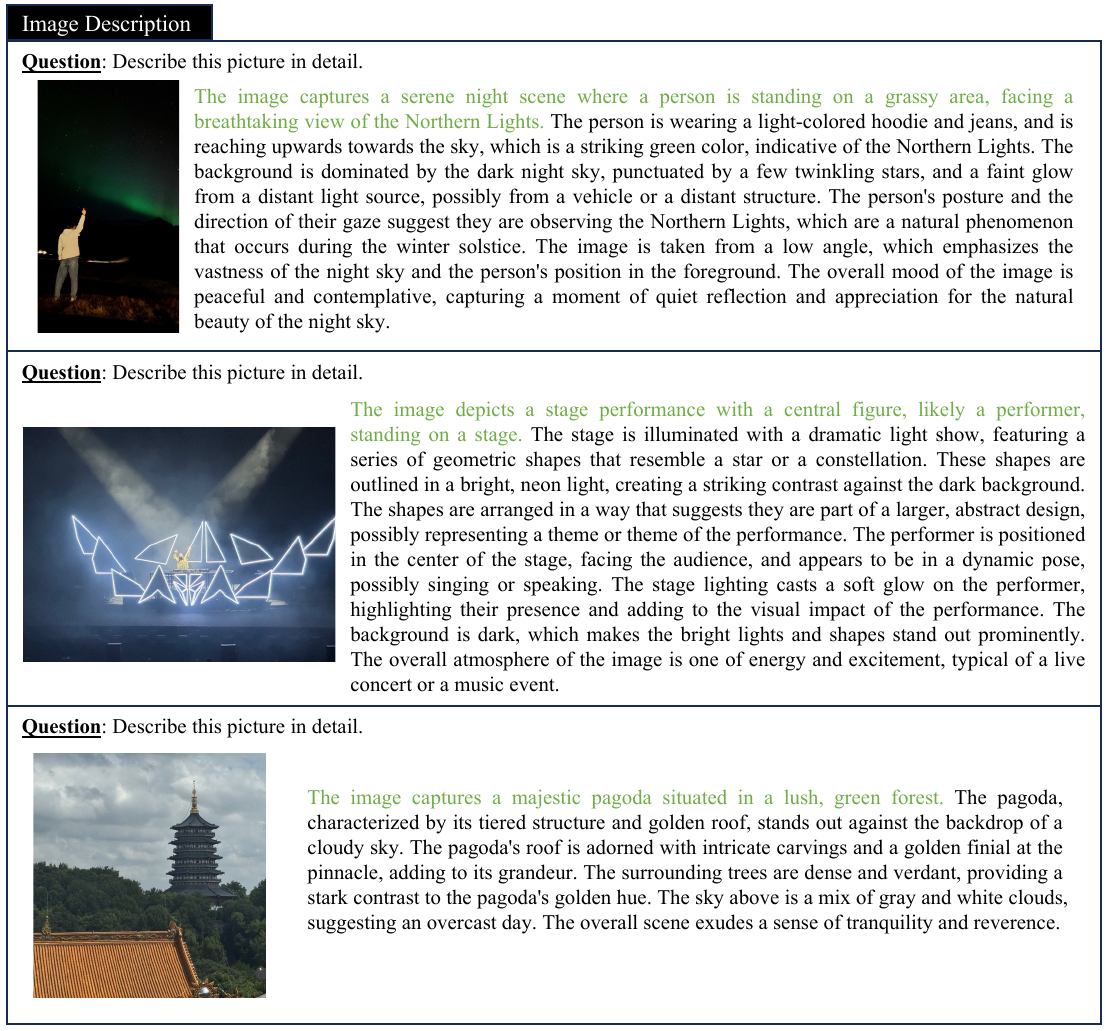}
\caption{Qualitative evaluation on real-world scenarios.}
\label{fig:real}
\end{figure*}

To further evaluate real-world performance, particularly on unseen visual content, Figure~\ref{fig:real}  showcases results on authentic, in-the-wild photographs. These images capture realistic deployment challenges, including variations in lighting, resolution, and scene composition. Despite these conditions, the 2-bit quantized model exhibits strong visual comprehension and generates accurate, contextually appropriate responses, demonstrating its robustness and readiness for practical applications in real-world settings.

\begin{table*}[ht]
  \centering
  \small
  \setlength{\tabcolsep}{0.6mm} 
      \begin{tabular}{cclccccccccc}
        \toprule
        Model & Bitwidth & \multicolumn{1}{c}{Method} & MMBench	& MMStar & MMMU &	Hallusion.	& AI2D	& OCRBench & SEED & ScienceQA  & Avg. ($\uparrow$) \\
        \midrule
        
        \multirow{10}{*}{InternVL2.5-1B} & BF16 & - & 67.38 &	50.73	& 38.11	& 39.22	& 69.14	& 77.6 & 70.66 & 93.04 & 63.23 \\
        \cmidrule(lr){2-12}
        ~ & \multirow{5}{*}{4-bit} & RTN & 61.30 &	48.80	& 36.78	& 40.03	& 65.93	& 72.9	& 68.89 & 89.08 & 60.46 \\
        ~ & ~ & GPTQ (ICLR'23) & 60.72 & 45.00 & 33.89 & 40.42 & 64.64 & 70.10 & 64.14 & 86.70 & 58.70 \\
        ~ & ~ & AWQ (MLSys'24) & 63.20 & 45.67 & 36.67 & \textbf{41.50} & 65.03 & 71.40 & 68.08 & 87.65 & 59.90 \\
        ~ & ~ & MBQ (CVPR'25) & 62.27&	47.4&	35.89&	40.79&	66.26&	75.6&	69.27&	88.79&	60.78 \\
        ~ & ~ & \textbf{SPEED-Q (Ours)$\ddagger$} & \textbf{66.41} &	\textbf{50.27} & \textbf{36.89} & 41.41 &	\textbf{68.36} & \textbf{75.8} & \textbf{71.47} & \textbf{89.08} &	\textbf{62.46} \\
        \cmidrule(lr){2-12}
        ~ & \multirow{4}{*}{2-bit} & RTN & 0.12 & 6.73 & 7.44 & 2.10 & 6.35	& 0.5 & 7.33 & 7.96 & 4.82 \\
        ~ & ~ & GPTQ (ICLR'23) & 0.00 & 10.2 & 9.67 & 1.82 & 10.59 & 0.4 & 9.03 & 12.73 & 6.81 \\
        ~ & ~ & MBQ (CVPR'25) & 0.12 & 7.8 & 10.0 & 0.94 & 8.52 & 0.6 & 8.26 & 9.82 & 5.76 \\
        ~ & ~ & \textbf{SPEED-Q (Ours)$\ddagger$} & \textbf{53.68}	& \textbf{43.6}	& \textbf{33.33}	& \textbf{37.04}	& \textbf{59.42}	& \textbf{61.2}	& \textbf{66.78} & 	\textbf{61.18}	& \textbf{52.03}\\
        \bottomrule
      \end{tabular}
  \caption{Main results on InternVL2.5 families, as an additional supplement to Table 1 in the main paper.}
  \label{supp-tab1}
\end{table*}

\section{Enabling Reasoning in Quantized VLMs}
\begin{figure*}[h]
\centering
\begin{tcolorbox}[colback=black!5!white,colframe=black!75!black,title=Reasoning Case1]
\includegraphics[width=5cm]{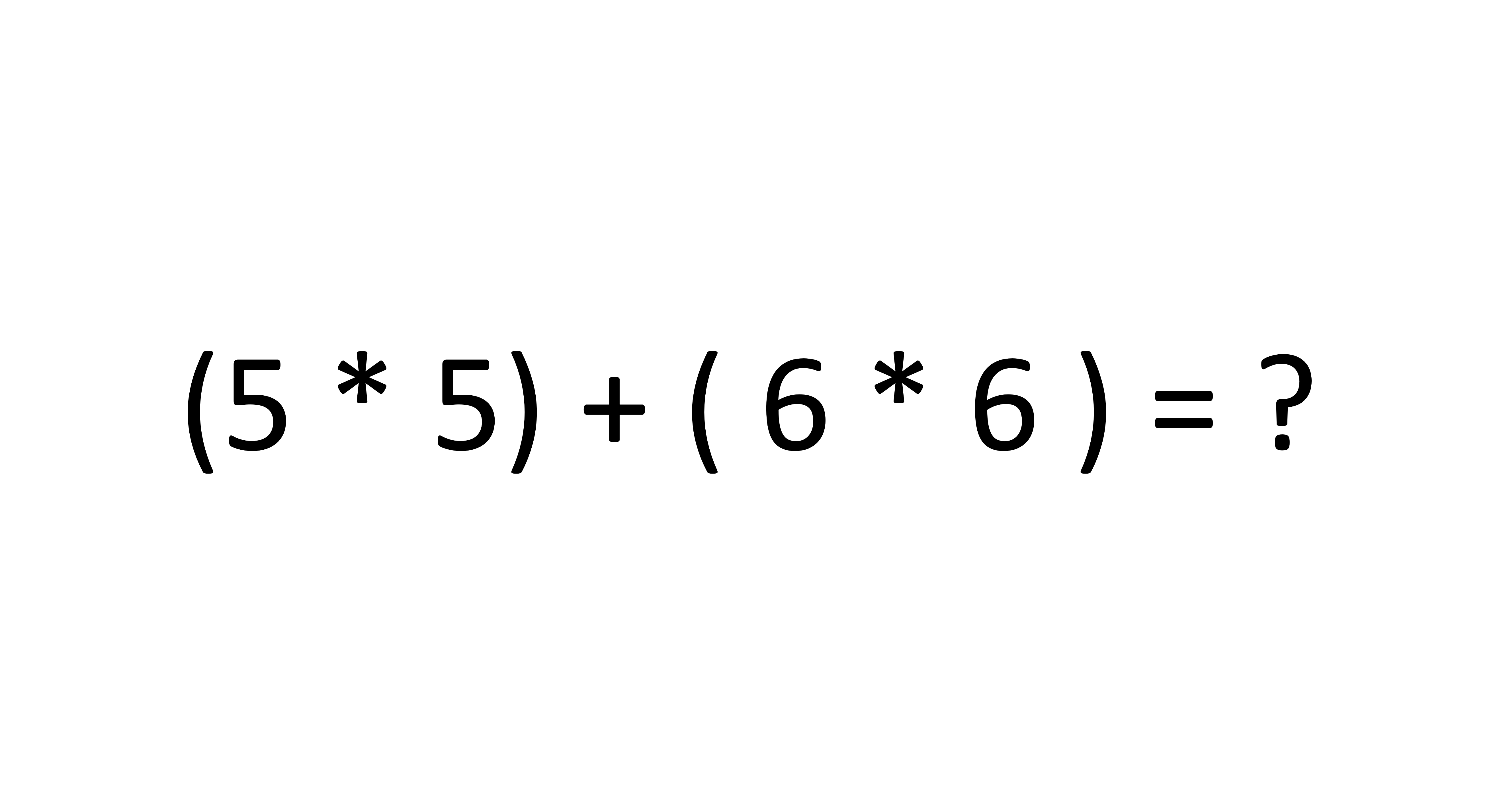}

What is the result of the formula in the figure?

\tcbsubtitle{Model Response}
$<$think$>$
\\
First, I need to calculate the multiplication parts of the expression: 5 multiplied by 5 and 6 multiplied by 6.
\\
\\
Next, I'll add the results of these multiplications together.
\\
\\
Finally, I'll present the result as the answer.
\\
\\
$<$/think$>$
\\
$<$answer$>$
\\
The result of the expression \((5 \times 5) + (6 \times 6)\) is calculated as follows:
\\
\\
1. Multiply 5 by 5: \(5 \times 5 = 25\) \\
2. Multiply 6 by 6: \(6 \times 6 = 36\) \\
3. Add the results: \(25 + 36 = 61\) \\
\\
\\
Therefore, the result of the formula is \(\boxed{61}\). \\
$<$/answer$>$
\end{tcolorbox}

\begin{tcolorbox}[colback=black!5!white,colframe=black!75!black,title=Reasoning Case2]
\includegraphics[width=5cm]{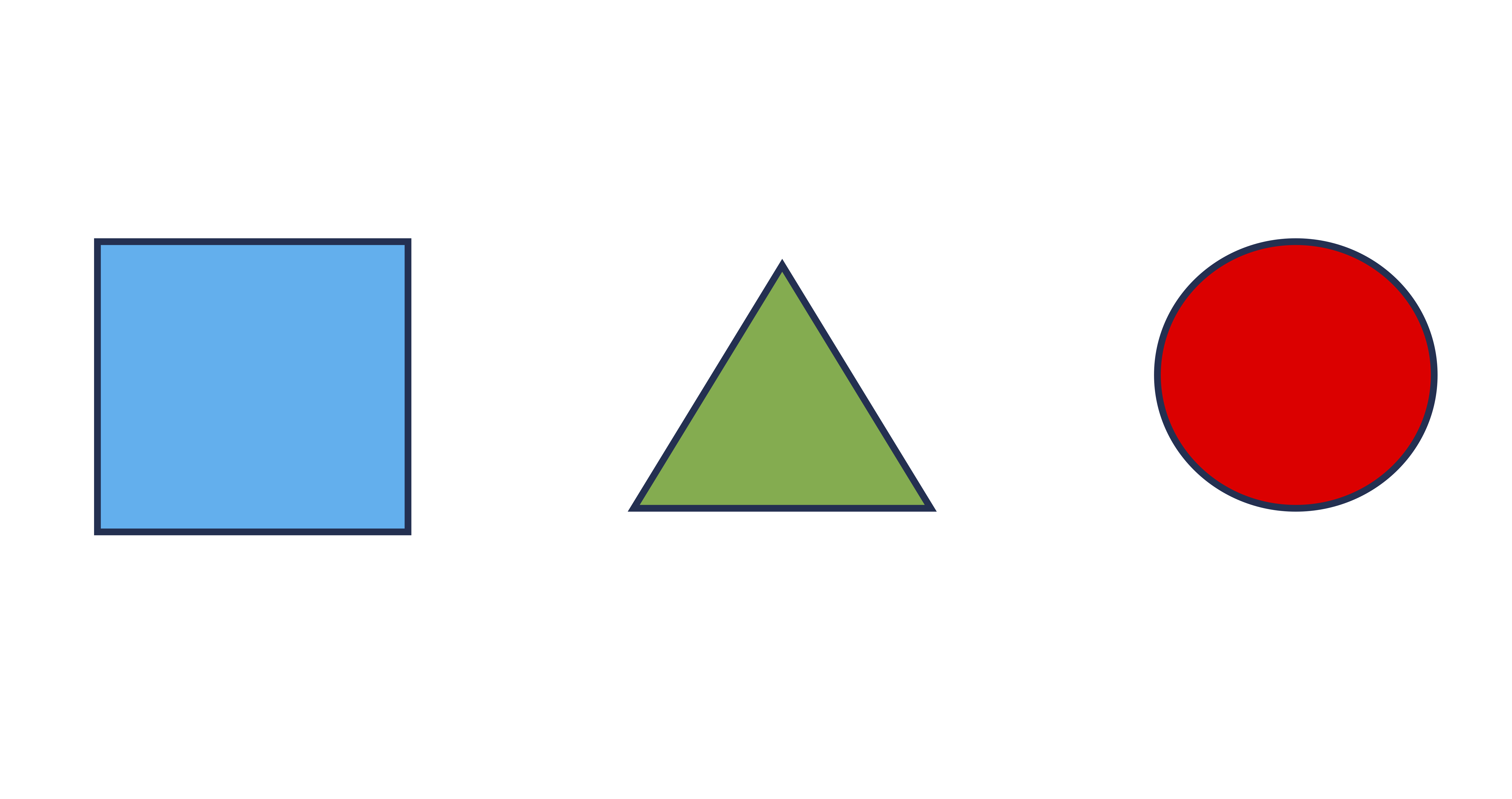}

What color is the triangle in the picture?

\tcbsubtitle{Model Response}
$<$think$>$
\\
\\
First, I need to identify the triangle in the image. The triangle is positioned in the middle row, which is the second object from the left.
\\
\\
Next, I will determine the color of the triangle. The triangle is described as having a green color.
\\
\\
Finally, I will present the answer clearly and concisely.
\\
$<$/think$>$
\\
$<$answer$>$
\\
The triangle is green.
\\
$<$/answer$>$

\end{tcolorbox}
\caption{Example of reasoning ability in InternVL3-1B-SPEED-Q (4-bit) via SFT.}
\label{fig:reason}
\end{figure*}

By applying reasoning-aware supervised fine-tuning (SFT) on the 4-bit quantized InternVL3-1B using the VLAA-Thinking dataset~\cite{vl-thinking2025}, the model demonstrates chain-of-thought reasoning capabilities, as shown in Figure~\ref{fig:reason}.
These reasoning skills are typically observed in large, full-precision models. Achieving such capabilities in a compact, low-bit vision-language model via efficient fine-tuning marks a significant advance, demonstrating that highly quantized models can support advanced reasoning. These findings highlight the practical potential of deploying intelligent, reasoning-enabled vision-language systems on resource-constrained edge devices.

\end{document}